%% file: paper.tex
\documentclass[10pt,twocolumn,letterpaper]{article}

\usepackage{cvpr}
\usepackage{times}
\usepackage{epsfig}
\usepackage{graphicx}
\usepackage{amsmath}
\usepackage{amssymb}
\usepackage{amsthm}
\usepackage{color}
\usepackage[table]{xcolor}
\usepackage[lofdepth,lotdepth, caption=false,font=footnotesize]{subfig}
\usepackage{hhline}
\usepackage{cite}
\usepackage{stfloats}
\definecolor{Gray}{gray}{0.85}
\input{figs}


\newcommand{\figref}[1]{Fig.~\ref{fig:#1}}

\newcommand{\secref}[1]{Section~\ref{sec:#1}}
\usepackage[pagebackref=true,breaklinks=true,letterpaper=true,colorlinks,bookmarks=false]{hyperref}

 \cvprfinalcopy % *** Uncomment this line for the final submission

\def\cvprPaperID{14} % *** Enter the CVPR Paper ID here

\ifcvprfinal\fi
\begin{document}

%%%%%%%%% TITLE
\title{Infant Contact-less Non-Nutritive Sucking Pattern Quantification\\ via Facial Gesture Analysis}

\author{Xiaofei Huang, Alaina Martens, Emily Zimmerman, and Sarah Ostadabbas$^*$\\
Northeastern University\\
Boston, MA, USA\\
{\tt\small ostadabbas@ece.neu.edu}
\thanks{Source code available at: \url{https://web.northeastern.edu/ostadabbas/software/}}
}

\maketitle
%\thispagestyle{empty}

%%%%%%%%%%%%%%%%%%%%% ABSTRACT %%%%%%%%%%%%%%%%%%%%%%%%%%%%%%%%%%%%%%%%%%%%%%%%%%%%%%%%
\begin{abstract}
Non-nutritive sucking (NNS) is defined as the sucking action that occurs when a finger, pacifier, or other object is placed in the baby's mouth, but there is no nutrient delivered. In addition to providing a sense of safety, NNS even can be regarded as an indicator of infant's central nervous system development. The rich data, such as sucking frequency, the number of cycles, and their amplitude during baby's non-nutritive sucking is important clue for judging the brain development of infants or preterm infants. Nowadays most researchers are collecting NNS data by using some contact devices such as pressure transducers. However, such invasive contact will have a direct impact on the baby's natural sucking behavior, resulting in significant distortion in the collected data. Therefore, we propose a novel contact-less NNS data acquisition and quantification scheme, which leverages the facial landmarks tracking technology to extract the movement signals of baby's jaw from recorded baby's sucking video. Since completion of the sucking action requires a large amount of synchronous coordination and neural integration of the facial muscles and the cranial nerves, the facial muscle movement signals accompanying baby's sucking pacifier can indirectly replace the NNS signal. We have evaluated our method on videos collected from several infants during their NNS behaviors and we have achieved the quantified NNS patterns closely comparable to results from visual inspection as well as contact-based sensor readings. 
\end{abstract}
\vspace{-.2in}
%%%%%%%%%%%%%%%%%%%% BODY TEXT %%%%%%%%%%%%%%%%%%%%%%%%%%%%%%%%%%%%%%%%%%%%%%%%%%%%%%%%%

\section{Introduction}
\subsection{Study motivation}
\label{sec:motiv}
Non-nutritive sucking (NNS) is one of the earliest motor behaviors that occur after a baby is born. It refers to the sucking action that occurs when a finger, pacifier, or other object is placed in the baby's mouth, but there is no nutrient delivered, as opposed to nutritive sucking, which is the sequence that occurs when fluid is being introduced \cite{foster2016non}. Typical NNS pattern (bursts of suck and pause periods for respiration alternate) is characterized by 6--12 suck cycles per burst with an intra-burst frequency of around two suck cycles per second ($\sim$2Hz) for normal young infants (from 4 days to 6 months) \cite{wolff1968serial}. 

It is known that the NNS behavior is an effective way for infants to seek self-comfort. Moreover, it promotes the development of neonatal sucking response and regulate the secretion of gastrointestinal hormones \cite{premji2000gastrointestinal}. It even can be regarded as an indicator of the progress in the infant's central nervous system development \cite{anthony2010value,bingham2009deprivation,poore2009suck}. The rhythmical properties of NNS could be an objective clinical clue for judging the effects of congenital abnormalities and perinatal stress on the brain function of the young infant \cite{wolff1968serial}. 

There is emerging evidence revealing that infant NNS and early feeding physiology is linked to oral feeding behaviors \cite{mccain1995promotion}, childhood language \cite{adams2013association, malas2015feeding, malas2017prior}, childhood motor abilities \cite{wolthuis2017sucking}, IQ \cite{wolthuis2017sucking} and overall
neurodevelopment \cite{wolthuis2017sucking,wolthuis2015association}. Associations from these previous studies were in the same direction: better NNS is linked to higher scores and improved behaviors. Because the human infant is born with very few motor capabilities, NNS study is one of the assessments that can be done early in development that may be predictive of future neurodevelopment.

Often, many researchers collect NNS data by using contact devices being inserted inside the baby's mouth that have embedded sensors such as pressure transducers \cite{lau2001quantitative,zimmerman2017patterned,wolff1968serial}. However, due to the high price of these measurement tools and being less friendly for non-professionals to operate with, home-use of these devices has been hindered and they are limited to the lab/clinic use only. Moreover, such invasive contact will have a direct impact on the baby's natural sucking behavior, resulting in significant distortion in the collected data. Besides, the collected sensor data are often only analyzed by visual inspection and manual burst counting rather than using automatic analytical methods that can be robustly applied on large sets of sensor data to assure the objectivity of the results. In addition, when sensors are not available, clinicians often assess suck using a gloved finger, which is a highly subjective process. Hence, there is a crucial need for a contact-less NNS data collection and analysis system, which is able to collect NNS signal in babies' natural settings, and extract parameters such as sucking cycles and their frequencies automatically. A vision-based NNS pattern quantification using advanced computer vision algorithms can offer a valuable means in studying the relationship among sucking difficulties, oromotor delays and subsequent neurodevelopment in early life of an infant, which allows for timely diagnosis and treatment planning if needed.

\subsection{Related Work}
The most commonly used NNS data acquisition device is a pacifier equipped with a pressure sensor. As described in \cite{zimmerman2017patterned}, the pressure transducer is utilized for the measurement system to detect and measure the infants' NNS patterns during infant sucking pacifier. The pressure transducer is housed within the pacifier handle or in a separate boxed container. INNARA HEALTH designed a NTrainer System \cite{poore2008patterned} to improve feeding development for premature and newborn infants by reinforcing NNS. Their NNS evaluation session uses the Actifier, which is a specially designed system that uses a Honeywell pressure transducer coupled to a custom Delrin receiver with a sterile soothie silicone pacifier to measure the force generated by the lips, tongue and jaw during sucking behaviour. However, this traditional NNS data collection method has some shortcomings. On one hand, the high acquisition cost is not suitable for a wide range of applications. On the other hand, contact-type acquisition may cause deformation in the infant's sucking pattern. The pacifier coupled with a pressure transducer will slightly displace the nipple making it slightly harder, causing a change in the feeling of sucking. Additionally, there is another rarely used sucking pattern observation method as mentioned in \cite{da2010sucking}: assessing infant's sucking pattern with non-invasive Neonatal Oral-Motor Assessment Scale (NOMAS) based on recorded feeding videos. The NOMAS demonstrates three sucking patterns: normal (or mature),  disorganised, and dysfunctional sucking patterns. Jaw movements and some tongue movements are scored as observed from the video recordings, and the other tongue movements are scored indirectly from the movements of lips, cheeks and the base of the mouth, as described in the NOMAS manual \cite{da2008reliability}.

In the fields of computer vision and computer graphics, extensive research exists in the areas of facial capture, tracking, animation, and recognition. Facial landmarks detection technology is no longer limited to face recognition, reconstruction, and identification, but more and more is applied to human behavioral research or reasoning about medical/psychological conditions with facial symptoms. In \cite{zeng2006spontaneous}, a 3D face tracker was used to detect the arbitrary behavior of the person in the natural setting. Both geometry and appearance features were extracted based on the 3D face model. A multimodal approach, combining facial movements and several peripheral physiological signals analysis was proposed in \cite{yin2018facial,yin2018affective} to decode individualized affective experiences under positive and negative emotional contexts. In \cite{wang2008automated}, author presented a computational framework that creates probabilistic expression profiles for video data and could potentially help to automatically quantify emotional expression differences between patients with neuropsychiatric disorders and healthy controls. 

\subsection{Our Contribution}
Due to the drawbacks of the current contact-based NNS data collection systems, this paper introduces a vision-based contact-less NNS data acquisition and quantification method, leveraging the recent advancements in the face recognition technologies. Our method automatically extracts movements of the baby's jaw's landmark in a video by tracking the 2D facial landmarks and then fitting a 3D morphable model (3DMM) to generate 3D facial landmarks. When a baby sucks his/her regular pacifier, the facial muscles and joints need to work in coordination, especially the regular movements at the mouth and lower jaw are closely related to the NNS pattern. Since the landmarks of the mouth are covered by the pacifier, we chose the jaw's landmark movement signal to replace the NNS signal, indirectly. Finally, the suck cycles and frequency of NNS pattern are calculated according to the landmark movement signal after denoising. In short, the main contributions of our novel NNS data collection and analysis approach are: (1) presenting a video-based contact-less data collection method for infant NNS pattern acquisition that can be adopted in babies' home without effecting their natural behaviors, and (2) developing a computer vision based toolbox to automatically process and analyze the infant suck video to quantify the important NNS features.

\section{Materials and Methods}
\subsection{Data Collection Procedure}
In this work, we collected two sets of data to validate our innovative approach. The first set of data includes 5 videos from different infants sucking a sensorized pacifier and their synchronous NNS data, which is traced by a pressure transducer, collected by the Speech and Neurodevelopment Lab of Northeastern University \figref{exampleA} shows a screenshot from a video that was recorded simultaneously while the sensory data was being collected.  The NNS samples were collected two weeks before or after the infant's 3 month birthday in their natural home environment. Parents are instructed to either cradle hold their infant or place their infant on their back. Infants were offered a Soothie pacifier (Philips Avent), which is connected to a custom measurement system to detect and measure the NNS pattern. The system utilizes a pressure transducer (Honeywell TruStabilityTM HSC Series Pressure Sensor, Morris Plains, NJ, USA) that measures the pressure within the pacifier during infant sucking. Researchers collected a 2$\sim$5 minutes sample depending on the infant's tolerance of the pacifier. 

\example
The second set of data (another 5 videos) were crawled from YouTube. These videos (an example is shown in \figref{exampleB}) were recorded in different environments when babies were lying down and sucking ordinary pacifiers. Because of the lack of video introduction about baby's age, current state, and video equipment, etc., we only speculated based on the videos that the age of these subjects is less than 6 months. The subjects was videoed in profile so that their jaws, the base of the mouth, lips and cheeks were clearly visible. And four of the videos were recorded when the babies were awake and the other was asleep. There are no other external factors or devices that interfere with the NNS behavior during the recording process. The average length of each video is about 20$\sim$30 seconds.

\mainMenu
A vision-based approach that replaces contact-based sensory data sources with non-contact video sources is introduced to trace infant's NNS patterns. In order to extract the NNS pattern by analyzing facial gesture, we designed a toolbox to track and analyze facial landmarks from videos. The interactive interface of facial gesture analysis toolbox is shown in \figref{mainMenu}. A previously-introduced 3D face landmark localization and tracking approach is employed as the first stage of our toolbox to track the movements of landmarks in videos. After landmarks tracking, the interface of landmarks processing, including particular landmark movement display and signal filtering can be utilized for facial gesture analysis. 

\subsection{Landmarks Tracking}
\tracking
As shown in \figref{tracking}, the Landmarks Tracking interface allows the user to choose an infant's sucking  video as the data source in order to automatically track and save the 68 3D facial landmarks positions, which are  enough to present the facial gesture. For the landmarks tracking, we detect face and track facial landmark locations following the method described in \cite{yin2018facial}: (1) localizing 2D facial landmarks for each frame of the infant's sucking video, by employing the 2D face alignment algorithm proposed in \cite{kazemi2014one}, and (2) as introduced in \cite{huber2016multiresolution}, estimating the 3D facial landmarks by fitting a 3D morphable face model to the 2D landmark locations, while decoupling the 3D head movement from the facial landmark movements.
\outputVideo

A cascade of regressors in \cite{kazemi2014one} is applied to precisely estimate the position of the 68 2D facial landmarks. Let the shape vector $S = (x^T_1,x^T_2,\dots,x^T_{68})^T \in \mathbb{R}^{2\times68}$ represents the coordinates of all the facial landmarks in an image $I$. $\hat{S}^{(t)}$ is the current estimate of $S$. In each level of cascade, estimated shape vector is refined by adding update vector predicted by the previous regression:
\begin{align}
    \hat{S}^{(t+1)} = \hat{S}^{(t)} + r_t(I,\hat{S}^{(t)}),
\end{align}
where $r_t(.,.)$ denotes regressor at time step $t$, and the ininital estimate $\hat{S}^{(0)}$ is defined as the mean shape of the training data centered and scaled according to the face bounding box, generated by a frontal face detector, which is made using the classic Histogram of Oriented Gradients (HOG) features combined with a linear classifier.
Each regressor in the cascade is learned via the gradient tree boosting algorithm with a squared error loss function. 

After localizing 2D facial landmarks for each frame, we estimate 3D landmark coordinates by fitting a low-resolution Surrey face model to the 2D landmark locations. This model includes a 3D Morphable Model (3DMM) and accompanying metadata, like a 2D texture representation and landmark annotations \cite{huber2016multiresolution}. 3DMM is generated in dense correspondence based on 3D meshes of faces. Assuming there are $N$ mesh vertices, a face can be represented by a three dimensional shape vector $S\in\mathbb{R}^{3N}$. 3DMM includes two principal component analysis (PCA) models, one for the shape, which could be used to reconstruct a 3D face from 2D image, and one for the color information.  Each PCA model, $(\bar{v},\sigma,V)$, consists of the mean components of mesh vertices $\bar{v}$, a set of principal components for all meshes $V = [v_1, \dots, v_K]$, and standard deviations for all components $\sigma$. $K$ is the number of principal components. Then the shape vector of novel face can be generated:
\begin{align}
    S_i  = \bar{v} + \sum_{k=1}^{K}\alpha_k\sigma_kv_k ,
\end{align}
where $\alpha$ are the 3D face instance coordinates in the PCA shape space.

As the landmark fitting algorithm introduced in \cite{aldrian2013inverse}, first the pose of the face is estimated by assuming an affine camera model and implementing the gold standard algorithm of Hartley \& Zisserman \cite{hartley2003multiple}, which finds a least squares approximation of a camera matrix given a number of 2D-3D point pairs. Then, the estimated 3$\times$4 affine camera matrix is used to eliminate the interference of head movement and compute homogeneous coordinates of the 3D feature points projected to 2D, which is one of the main factors of cost function. By minimizing the cost function, we can find the most likely vector of PCA shape coefficients for 3D landmark coordinates. The pose estimation and shape fitting steps can be iterated if desired to refine the estimates. Finally, this camera matrix represents the 3D pose of head. The head movement is then composed by successive poses of head. According to the frontalised 3D shape coefficients, the 3D facial landmarks that are decoupled from the 3D head movement are then generated.

When the 3D landmarks tracking of the video is complete, the toolbox produces 68 3D frontalised facial landmark coordinates and an output video that is synchronized with the original video. The snapshot of output video is shown in \figref{outputVideo}.

\subsection{Landmarks Processing}
\processing
After the 3D facial landmarks for each frame of video are generated, the second stage of our toolbox, Landmarks Processing, can be used to display output video, visualize movement of a specific landmark, and denoise the raw data to analyze landmark movement and quantify NNS pattern. The landmarks processing interface, shown in \figref{processing}, provides a number of functions that make it easy for the users to analyze facial gesture and figure out its association with the NNS patterns. By loading the 3D landmarks file, generated at the previous stage, the corresponding output video (refers to \figref{outputVideo}) could be shown in a pop-up window. The pop-up visualization of 68 facial landmarks annotations allows users to lightly obtain the index of a landmark they want to observe. The raw and filtered movement signal of designated landmark are displayed on top and bottom canvases, respectively. Users are able to set the parameters for the denoising filter. By pressing ``Apply Bandpass Filter'' button, the filter will be applied to all of the movement signals shown in top canvas. 

The location of each 3D landmark in first frame is regarded as the initial position. Movement signal of a landmark is the variation in coordinates over image sequence by computing the distance/displacement between initial position and current position of this landmark. In order to comprehensively analyzing landmark movements, as you can see in \figref{outputVideo} and \figref{processing}, we considered three types of distance: Euclidean distance, horizontal displacement, and vertical displacement. The changes in these distances directly or indirectly reflect the deformation of face.

\subsection{NNS Pattern Quantification}
As we mentioned in the \secref{motiv}, the intra-burst frequency of NNS pattern is about 2Hz, that is, a baby has about two sucking actions in one second during a burst. For the landmark movement extracted from the video, since the frame rate of video is 30fps, and the landmark location of each frame is independently estimated, the deviation of predicted landmark position can be regarded as high frequency noise. The infrequent baby's head movements and  the hand-held camera movements can also be regarded as a low frequency disturbance. To eliminate these sources of noise, we provide a bandpass Butterworth filter to process the raw landmark movement signals. In order to better observe and study the relationship between the effective movement of landmark and the NNS pattern, the recommended setting for the low and high cutoff frequencies of the filter to be 0.3Hz and 3Hz, respectively. 

In the filtered signal, we define a peak greater than the average displacement as a suck cycle, and an interval with a continuous peak occurrence as a burst. Accordingly, the number of NNS cycles in each burst, the number of bursts in a given time interval, and each burst duration can  automatically be calculated from our peak detection algorithm. We can then calculate the average sucking frequency using these features as the total cycles number of all burst over the total duration of all burst. This series of NNS pattern parameters is generated by clicking ``Pattern Quantification'' button in the landmarks processing window as shown in \figref{processing}.

\section{Experimental Results}
We performed the facial gesture analysis as described above for the two sets of videos provided by the Speech and Neurodevelopment Lab and YouTube, and then compared the results with sensory signals and visual inspection, respectively. In addition, the NNS pattern was observed visually for rhythmic pacifier movement in each video. The start time and end time of each burst, and the sucking cycles of intra-burst were recorded manually. These recordings of visually observed and marked sucking movements as groundtruth were compared thereafter with our tracked landmark movement. Since the differences in stiffness and elasticity of pacifiers and the variability in strength of infants' sucking influenced the pressure data, the comparisons of absolute values for pressure were not meaningful. Therefore, at this stage of our study we only discuss the consistency of the rhythm.

\compLab
For the first validation set, sensory data acquisition was completed using the ADInstruments PowerLab and Labchart Pro software  used to analyze NNS dynamics. These suck dynamics include NNS burst duration (seconds), NNS cycles per burst, NNS cycles per minute, NNS bursts per minute, NNS frequency (Hz) and NNS amplitude (cmH2O). The degree of correspondence between the landmark movement generated by our toolbox and the raw sensory signal was assessed from an analysis using those sections of the recording where infant's facial landmarks could be tracked clearly. Therefore, we took a clip with high quality landmarks tracking in each video as our analysis object and process it to extract NNS pattern parameters. For illustration, a tracked landmark movement of a sample video is displayed at the top of \figref{compLab}, while the corresponding groundtruth of NNS pattern is in the bottom graph. Note that intervals containing more than 5 cycles are considered a burst. Obviously, by using our toolbox we detected three bursts, but only two actually were reported by the contact-based sensor. As we discussed in the \secref{motiv}, transducer affects the stiffness of nipple and thus the facial muscle movement. On the other hand, the occlusion of jaw's profile by the sensorized pacifier results in inaccurate landmark localization. These factors cause some of low peaks in the landmark movement signal to be ignored, thereby more bursts are split. Additionally, because of filtering process, the landmark movement lagged slightly behind the manual marking in general. 

\compYouTube
When testing the second set of data, YouTube videos, we obtained much more satisfied NNS patterns. A close concordance between a facial landmark movement and the manual marking of sucking movements was observed (see \figref{compYouTube}). As the amplitude of facial muscle movement of the baby when sucking the regular pacifier is greater than that of the pacifier coupled to a transducer, it is easier to extract pattern from a landmark movement with high signal-to-noise ratio. This also indirectly proves that our contact-less approach is superior to sensor-based detection.

\section{Conclusion and Future Work}
The present work aims to develop a novel and unobtrusive method for infant NNS pattern recording and analysis. We have designed a contact-less NNS pattern quantification scheme, which leverages the facial landmarks tracking technologies to trace the movement signals of baby's jaw based on the recorded baby's sucking video and then extract the parameters of sucking pattern from the landmark movement signal. Our experimental research is still at a primary stage. Due to the limited number of experimental subjects and the inconsistency between the recorded videos by the parents, the experimental results obtained above can only qualitatively establish the association between facial landmark movement and NNS pattern. In order to more accurately extract and define NNS patterns from facial landmark movements, we need more in-depth research to improve our current approach in the following proposed ways: (1) Since both 3DMM utilized in 3D landmark fitting or pre-trained model used by the CNN-based landmark tracker are generated based on the proportion of the adult's face size, this fact greatly affects the accuracy of infant facial landmarks tracking. Therefore, it is crucial to fine-tune a facial model on a sample infant dataset. (2) Under the premise that facial landmark localization is more accurate, we can quantitatively study the proportional relationship between amplitude of landmark movement and the infant sucking intensity.

In short, for NNS pattern recognition, although our acquisition and analysis method still need to be improved in reliability, it is undeniable that this is a new breakthrough in contrast to traditional contact-based collection methods. The objective and accurate extraction of the NNS pattern parameters without any external intrusion or interference with the state of pacifier is extremely important for linking infant sucking and feeding behaviors with subsequent speech-language production and cognition.

{\small
\bibliographystyle{ieee}
\bibliography{paper}
}
\end{document}

%% file: figs.tex
\newcommand{\example}{
\begin{figure}[t]
    \centering
    \subfloat[]{\label{fig:exampleA}\includegraphics[width=0.3\linewidth]{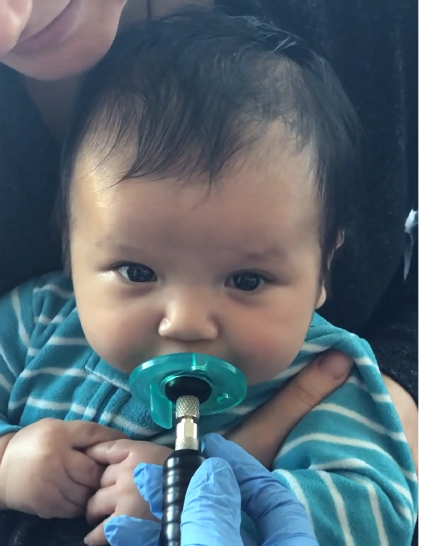}}
    \subfloat[]{\label{fig:exampleB}\includegraphics[width=0.51\linewidth]{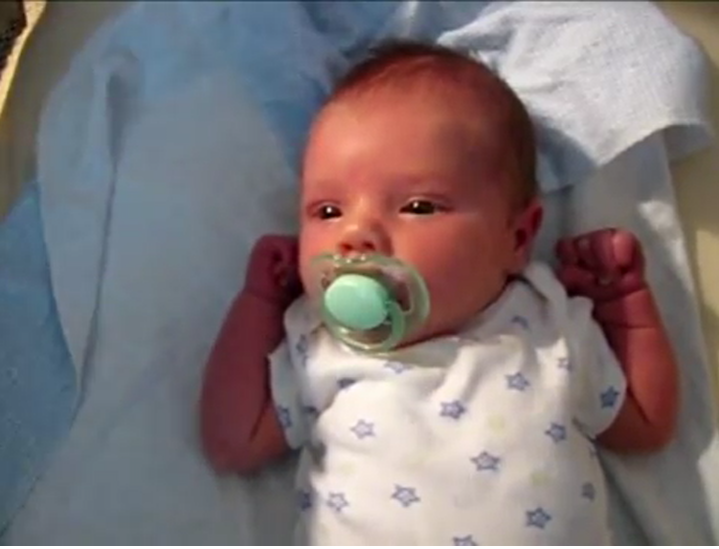}}
    \caption{A snapshot of a collected video from (a) 95-day-old infant using the sensorized pacifier at home, (b) a lying baby while sucking a regular pacifier.}
    \label{fig:example}
\end{figure}
}

\newcommand{\mainMenu}{
\begin{figure}[b]
    \centering
    \includegraphics[width=0.9\linewidth]{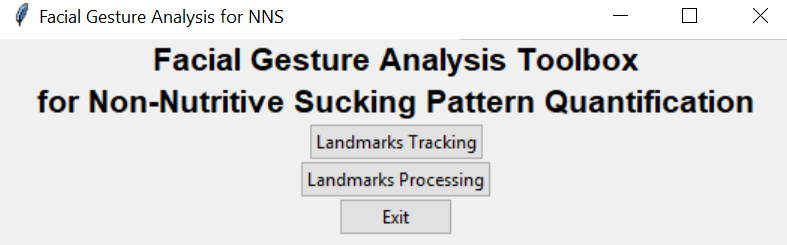}
    \caption{\footnotesize Opening menu of the facial gesture analysis toolbox for NNS pattern quantification.}    \label{fig:mainMenu}
\end{figure}
}

\newcommand{\tracking}{
\begin{figure}[b]
    \centering
    \includegraphics[width=0.9\linewidth]{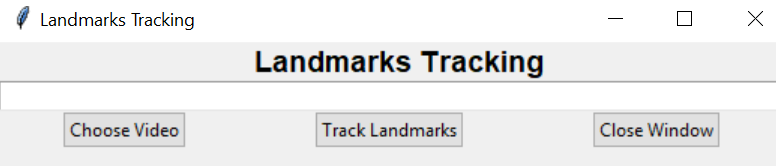}
    \caption{ \footnotesize Interface of Step I: Landmarks Tracking.}
    \label{fig:tracking}
\end{figure}
}

\newcommand{\outputVideo}{
\begin{figure}[t]
    \centering
    \includegraphics[width=0.9\linewidth]{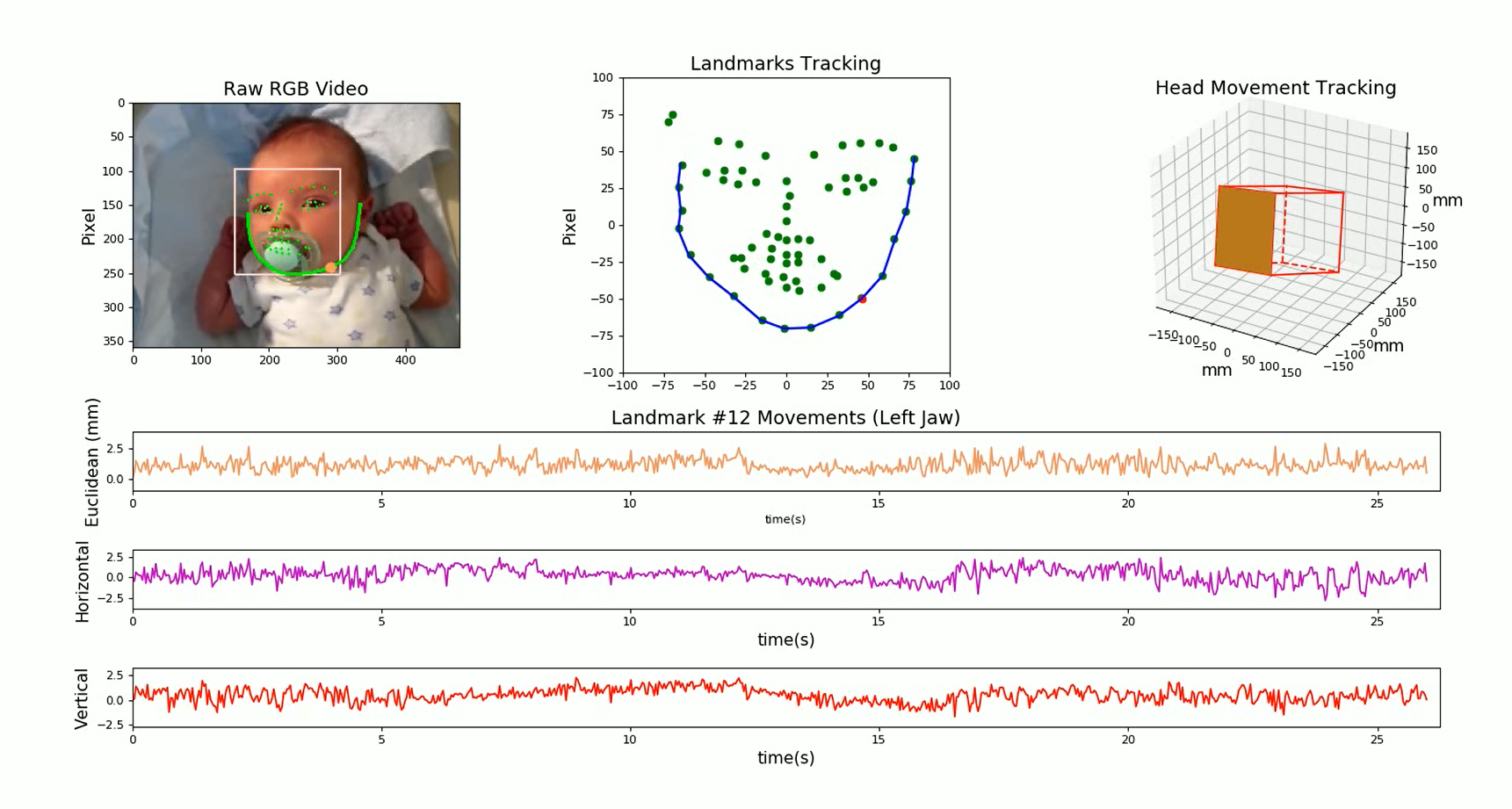}
    \caption{\footnotesize The snapshot of output video after landmarks tracking: the original video with 2D landmarks (left-top plot), the two-dimension trace of 3D landmarks tracking (mid-top plot), and 3D head movement tracking (right-top plot. The three bottom plots display the waveform of euclidean, horizontal and vertical movements respectively for one point of left jaw (landmark \#9), which is highlighted in left-top and mid-top plots.}
    \label{fig:outputVideo}
\end{figure}
}

\newcommand{\processing}{
\begin{figure}[b]
    \centering
    \includegraphics[width=0.9\linewidth]{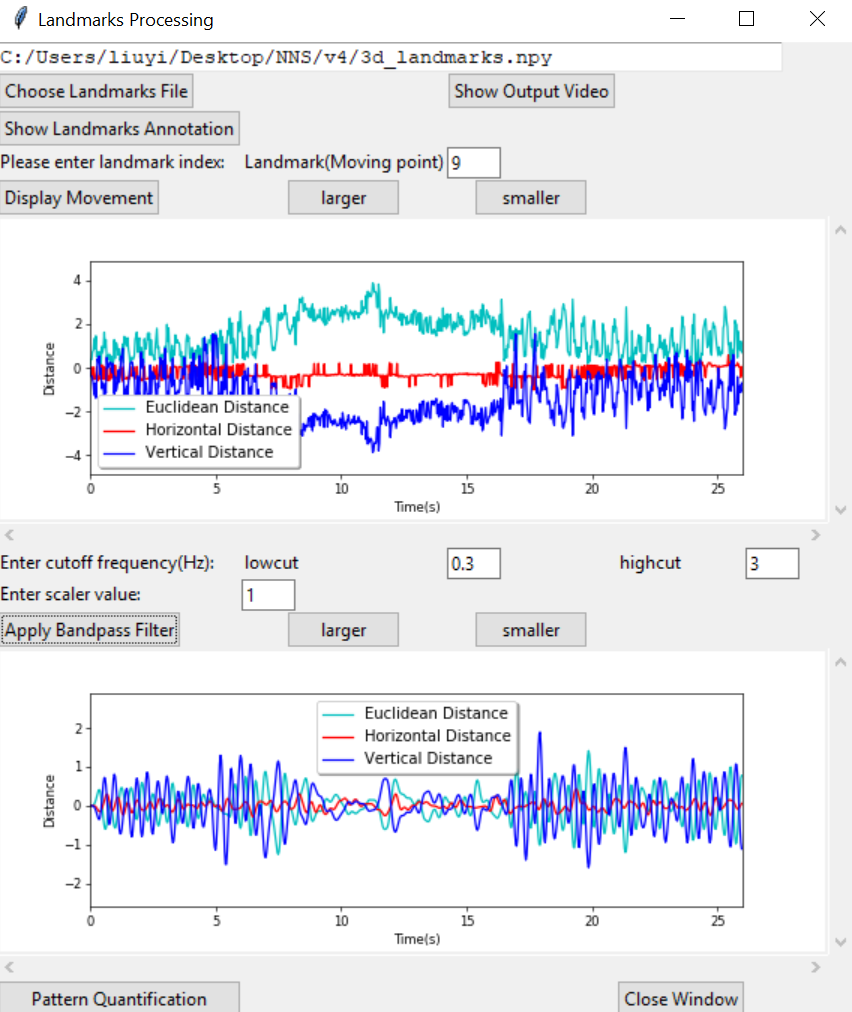}
    \caption{\footnotesize  Layout of the landmarks processing window.}
    \label{fig:processing}
\end{figure}
}

\newcommand{\compLab}{
\begin{figure}[b]
    \centering
    \includegraphics[width=0.9\linewidth]{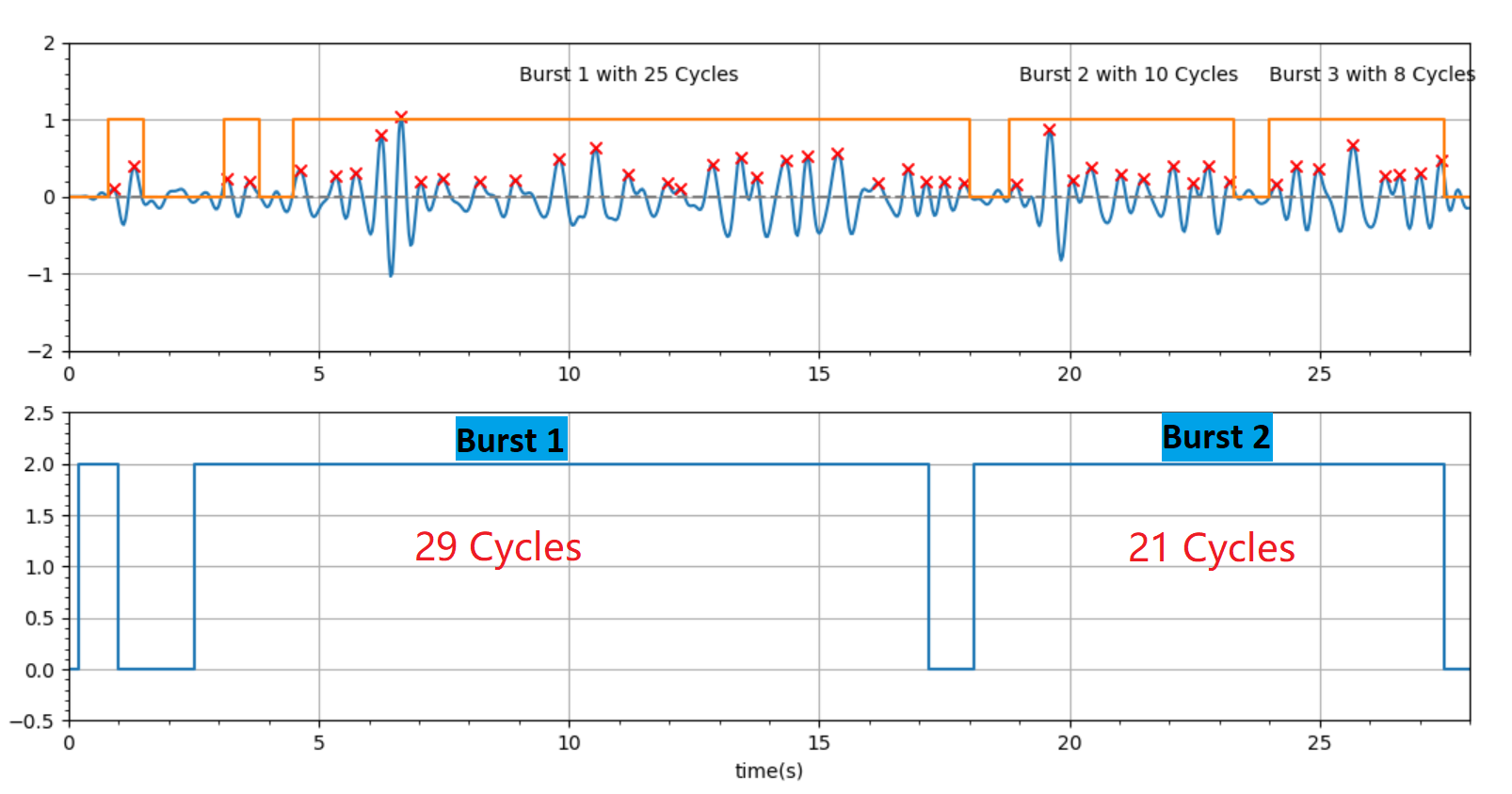}
    \caption{\footnotesize Video-based jaw's vertical movement detected by our toolbox (top plot) compared with the contact-based sensor readings inside a sensorized pacifier (bottom plot) in one of the lab's experimental videos. (Notice: here we ignored amplitude and only considered the frequency features).}
    \label{fig:compLab}
\end{figure}
}

\newcommand{\compYouTube}{
\begin{figure}[b]
    \centering
    \includegraphics[width=0.9\linewidth]{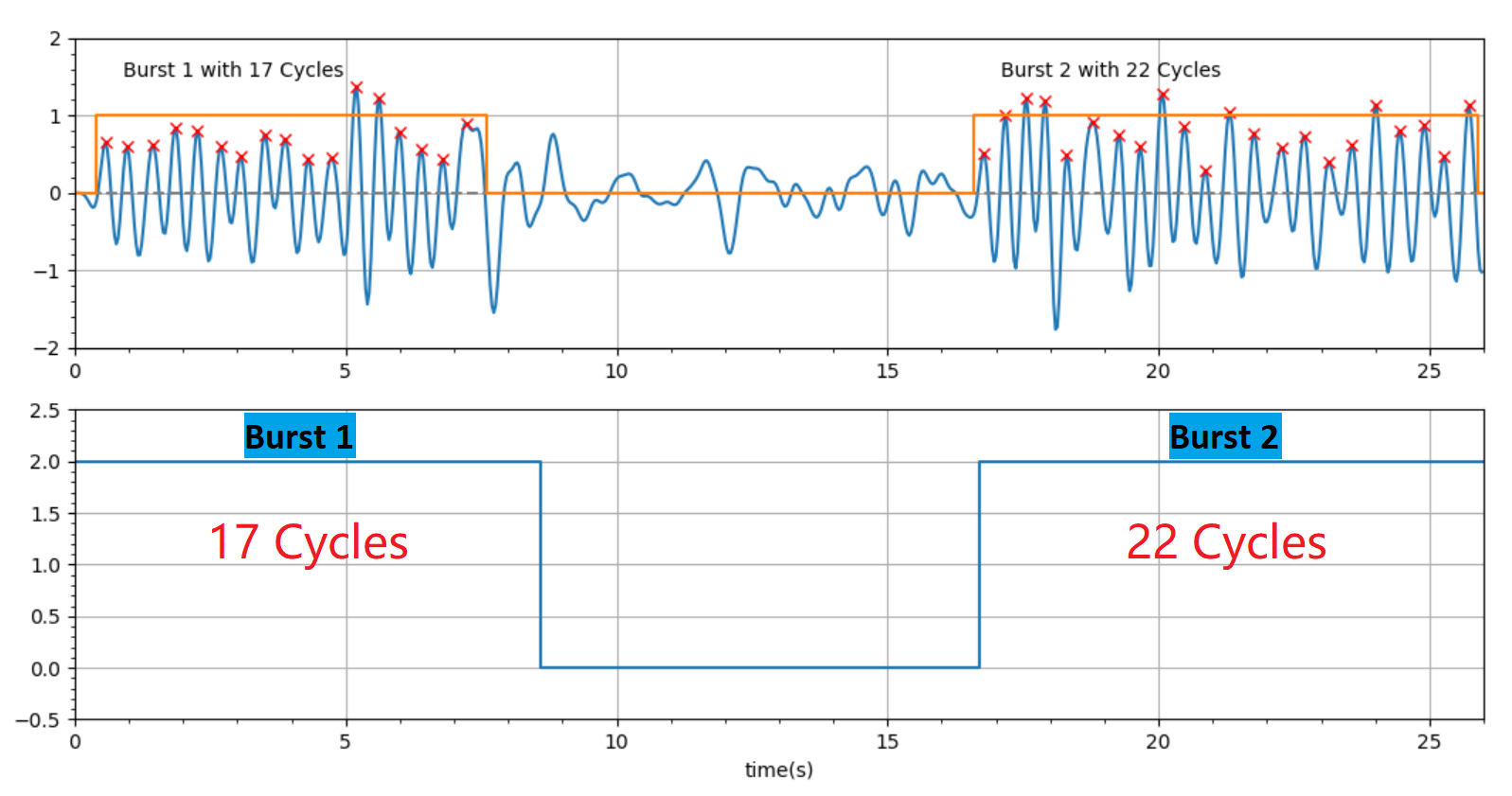}
    \caption{\footnotesize Video-based jaw's vertical movement detected by our toolbox (top plot) compared with manual marking of sucking movement (bottom plot) in one of the  YouTube videos.}
    \label{fig:compYouTube}
\end{figure}
}